\documentclass[a4paper,fleqn]{cas-sc}

\usepackage[authoryear,longnamesfirst]{natbib}
\usepackage{multicol}
\usepackage{graphicx}
\usepackage{subcaption}

\def\tsc#1{\csdef{#1}{\textsc{\lowercase{#1}}\xspace}}
\tsc{WGM}
\tsc{QE}

\begin{document}
\let\WriteBookmarks\relax
\def\floatpagepagefraction{1}
\def\textpagefraction{.001}

\shorttitle{Enhanced Forecasting for Logistic Hubs}    


\title [mode = title]{Enhanced Parcel Arrival Forecasting for Logistic Hubs: An Ensemble Deep Learning Approach}  


%

\author[1]{Xinyue Pan}

\cormark[1]


\ead{xpan67@gatech.edu}


\credit{Methodology, Software, Investigation, Formal analysis, Validation
Writing - Original Draft, Writing - Review \& Editing, Validation, Visualization}

\affiliation[1]{organization={Georgia Institute of Technology},
            addressline={755 Ferst Dr NW}, 
            city={Atlanta},
            postcode={30318}, 
            state={GA},
            country={USA}}

\author[1]{Yujia Xu}

\ead{yujia.xu@gatech.edu}
\credit{Methodology, Software, Investigation, Formal analysis, Writing - Original Draft}

\author[1]{Benoit Montreuil}
\ead{Benoit.Montreuil@isye.gatech.edu}
\credit{Conceptualization, Writing - Review \& Editing, Supervision, Project administration}

\cortext[1]{Corresponding author}

\fntext[1]{}

\nonumnote{}

\begin{abstract}
The rapid expansion of online shopping has increased the demand for timely parcel delivery, compelling logistics service providers to enhance the efficiency, agility, and predictability of their hub networks. In order to solve the problem, we propose a novel deep learning-based ensemble framework that leverages historical arrival patterns and real-time parcel status updates to forecast upcoming workloads at logistic hubs. This approach not only facilitates the generation of short-term forecasts, but also improves the accuracy of future hub workload predictions for more strategic planning and resource management. Empirical tests of the algorithm, conducted through a case study of a major city's parcel logistics, demonstrate the ensemble method's superiority over both traditional forecasting techniques and standalone deep learning models. Our findings highlight the significant potential of this method to improve operational efficiency in logistics hubs and advocate for its broader adoption.

\end{abstract}


\begin{highlights}
\item This study introduces a cutting-edge ensemble method that leverages deep learning to synthesize historical arrival patterns and real-time parcel status updates, significantly enhancing forecast accuracy for logistic hub workloads.
\item  Through rigorous empirical experimentation within a megacity parcel logistics scenario, the paper validates the superiority of the proposed model, demonstrating its ability to outperform both traditional forecasting methods and standalone advanced deep learning models.
\item  The findings highlight the potential of using sophisticated ensemble methods to reduce uncertainty and improve resource management, supporting the adoption of these advanced techniques for operational enhancements in logistics hubs.
\end{highlights}

\begin{keywords}
 \sep Deep Learning \sep Logistics Forecasting \sep Ensemble Methods \sep Real-time Data Processing \sep Supply Chain Management \sep Predictive Analytics
\end{keywords}

\maketitle

\section{Introduction}
In the past decade, the internet has transformed the retail and consumer industry by introducing a new shopping channel. For example, in the Chinese e-commerce market, the transaction volume reached about 4 trillion USD in 2018, according to the China Federation of Logistics and Purchasing (2019). \cite{Viu-Roig2020The} showed that global e-commerce sales significantly increased during the COVID-19 pandemic, highlighting rapid sector expansion . This surge in online shopping has increased the demand for parcel delivery, putting considerable pressure on logistic hubs to handle higher workloads efficiently. As parcel volumes grow, logistics service providers are also challenged by customer expectations for timely deliveries—a crucial factor in customer satisfaction and retention. Delays can result in more complaints, higher operational costs, and potential client loss. Additionally, the logistics competitive landscape is intensifying, with a notable rise in third-party logistics companies, according to IBISWorld. Thus, logistics hub operators are compelled to seek more effective delivery solutions to manage increasing demands and stay competitive in a rapidly evolving market.

Demand forecasting for parcel arrivals is crucial  as mentioned in \cite{pan2021assessing}, as it allows logistics hub operators to allocate resources effectively, maximizing utilization and throughput. Besides providing a foundation for decisions on infrastructure construction, equipment investment, and logistics development, accurate parcel arrival forecasts offer practical recommendations and theoretical support to logistics service providers (LSPs) \cite{tang2022forecasting}. These forecasts help optimize staff planning, minimize costs, enhance delivery efficiency, and improve overall competitiveness \cite{huang2019data}. Therefore, accurate demand forecasting for parcel arrivals is essential for efficient and effective logistics operations, ultimately contributing to business success in the logistics sector.

Previous studies have focused more on forecasting mid- to long-term daily arrival demand for example \cite{lin2013tourism, song2019review, ghalehkhondabi2019review}. However, short-term arrival demand prediction allows for flexible short-term plan adjustments and quick responses to volume changes. Moreover, many forecasting studies rely solely on past information, such as \cite{pai2014tourism, hess2021real}. However, history does not always repeat itself; for example, the COVID-19 outbreak caused an unexpected spike in online sales. Thus, a dynamic data-driven method to forecast short-term multi-step parcel arrival volumes was proposed to fill this research gap. This methodology, which is not constrained by forecast horizons or data time-frequency, allows for forecasting arrivals at every hub (node) and path (arc) in the logistics network. 

This research makes several significant contributions to the field of logistics forecasting by introducing innovative techniques and demonstrating their practicality. The contributions are as follows:

\begin{enumerate}[a]
    \item We have developed a novel ensemble forecasting method that dynamically integrates both historical arrival patterns and real-time updates on parcel status. This method enhances prediction accuracy by effectively utilizing up-to-date information, thereby reducing uncertainty and facilitating more effective planning and resource management within logistics hubs.
    
    \item Our research pioneers the application of advanced deep learning models in the logistics domain, specifically in the context of parcel arrival forecasting. We demonstrate how deep learning models can be adapted to significantly improve forecasting accuracy over traditional models, thus addressing a critical research gap and setting the stage for future applications in this area.
    
    \item The effectiveness of our proposed method was rigorously tested using a dataset over a 30-day period, which, despite its limited size, allowed for a comprehensive evaluation under constrained conditions. The promising results obtained from this experiment not only validate the robustness of our method but also underscore its potential scalability and adaptability to larger, more complex datasets.
\end{enumerate}

These contributions highlight the impact of our research on enhancing the operational efficiencies of logistics hubs through the application of cutting-edge forecasting techniques. Our work not only advances the theoretical framework of parcel logistics forecasting but also provides practical insights that could be instrumental in transforming current practices.
The rest of this work is structured as follows. Section 2 reviews literature relevant to demand forecasting in the parcel logistics industry. Our approaches, including the ensemble framework, arrival time prediction, and parcel demand forecasting, are covered in Section 3. The details and findings of computational experiments are described in Section 4 and 5. Finally, in Section 6, we discuss the directions for further research.

\section{Literature Review}
There have been limited studies on demand forecasting for operational scheduling in parcel logistics hubs. Numerous relevant studies have utilized incoming workload as input, often relying on approximate estimations based on historical data, strong assumptions, or random data generation, as mentioned by \cite{buckley2021hyperconnected}. For instance, \cite{tsui1992optimal} constructed a bi-linear optimization model for parcel logistics hubs using historical demand data for arriving parcels. Similarly, \cite{jarrah2016destination} employed the average monthly arrival workload as input for the destination-loader-door assignment problem. In another approach, \cite{bozer2008optimizing} presented a simulated annealing heuristic aimed at decreasing the material handling burden. Concurrently, \cite{haneyah2014throughput} focused on scheduling inbound containers to enhance sorter system throughput, with both studies assuming complete knowledge of the incoming workload. Additionally, \cite{fedtke2017layout} and \cite{boysen2010cross} conducted research on fully automated parcel logistics hubs using simulations. They employed random distributions for assigning destinations to arriving parcels.Although these forecasting approaches have significantly advanced operational decisions in parcel logistics hubs, there remains substantial potential for further development.

Recent advancements in real-time demand forecasting methods have demonstrated the potential of integrating classical forecasting with machine learning techniques. This hybrid approach enhances the responsiveness and accuracy of forecasting models, providing substantial improvements over traditional methodologies used in parcel logistics and other dynamic operational environments. Similar to the work done by \cite{hess2021real}, another compelling example of hybrid forecasting is presented by \cite{xu2019forecasting}, where a novel model combining SARIMA and SVR has been used in the aviation industry to forecast statistical indicators effectively. This model, particularly the SARIMA\_SVR3 variant, achieved notable accuracy, illustrating the benefits of incorporating Gaussian White Noise into the forecasting process.

The "bullwhip effect," which illustrates the increasing unpredictability in demand up the supply chain, also manifests in the complex logistics environment. As discussed by \cite{chen2000quantifying} and \cite{lee1997bullwhip}, sharing information from downstream to upstream entities can mitigate this issue and improve forecasting accuracy. In our context, if parcel logistics hubs within the same network collaborate, and downstream hubs supply upstream hubs with real-time information on arriving parcels, the logistics network can manage demand variation more effectively and reduce forecasting errors. The concept of data sharing among different supply chain entities has been advanced under the Physical Internet (PI) paradigm, as demonstrated by \cite{montreuil2011toward}. PI fosters cooperation across logistics hubs within the same network, enabling the more efficient establishment of cloud-based IT solutions necessary for data sharing. Additionally, \cite{pan2021data} and \cite{pan2021digital} demonstrated that PI is a crucial driver for digitalizing LSCM, as it provides a practical paradigm for digital supply chain architecture. Digitalizing LSCM facilitates data sharing more effectively than before. \cite{huang2019data} mentioned that with advancements in information technology, data can be continuously transferred to the hubs, enabling collaboration, rapid adaptation to unforeseen events, and improved supply chain efficiency and responsiveness. Previous research has demonstrated the feasibility of enhancing supply chain agility through live shared information, but few studies propose practical methods for realizing this potential.

Our investigation focuses on two distinct types of parcels. The parcel types are defined as follows: at the given observation time, type I parcels had not been ordered, whereas type II parcels had already been ordered and entered the logistics network. Both types of parcels will arrive at the target hub, requiring distinct approaches for each. Since type I parcels have not entered the network and their information is unknown, historical patterns will be used to forecast their arrival volumes. For type II parcels already in the network, live status information can be used to improve the forecast. Since the volume of type II parcels is known, the remaining task is to determine their arrival times at the target hub. For type II parcels, the focus will be on dynamic arrival time forecasts that incorporate updated parcel status. The next step is to convert predicted arrival times into expected parcel volumes. After obtaining arrival volumes for both ordered and unordered parcels, an ensemble model will be introduced to generate total arrival volume forecasts.

Many studies have been conducted on predicting travel times, but most have focused on automobiles on highways or buses. \cite{ahmed1979analysis} explored techniques for analyzing highway traffic volume and occupancy time series, determining that the autoregressive integrated moving-average (ARIMA) model provides the best results. \cite{Zhang2001ShortTermTT, Zhang2003ShorttermTT} proposed a linear model for estimating highway travel times, where coefficients vary as smooth functions of departure time. Similarly, \cite{Rice2004ASA} revealed a linear relationship between future and current travel times and consequently adopted a linear regression method with time-varying coefficients. As information technology advances and more data is collected, scientists increasingly find that machine learning and deep learning tools are beneficial for predicting travel time. \cite{wu2004travel} used support vector regression (SVM) to forecast automobile travel time using actual highway traffic data. Additionally, \cite{Bin2006BusAT} used SVM to forecast bus travel time. The travel time for a bus on the highway is more comparable to parcel travel time. In the urban network context, bus travel times should account for connections, waiting at stops, and delays at junctions. In the logistics network framework, we need to consider not only travel time between hubs but also dwell time, such as loading times and delays.

Tree-based models like Random Forest (RF) have become very popular because they are robust to noise, handle large datasets efficiently, and can capture non-linear relationships without requiring extensive data preprocessing, as shown by \cite{dietterich2002ensemble}. \cite{friedman2003multiple} mentioned that RF improves regression accuracy without significantly increasing computational complexity and reduces overfitting due to the random selection of features and training subsets. RF has been widely applied to bus travel time prediction, for example, by \cite{yu2018prediction, gal2017traveling, moreira2008travel}, but none of these studies focus on predicting travel time within a parcel logistics network. This is a significant gap, as compared to bus arrival times, parcel delivery times are more complex and difficult to forecast because buses typically follow predetermined routes and schedules. In contrast, parcels may have varying routes and intermediate stops, making their travel times more unpredictable. In this study, we bridge this gap by presenting an RF-based model to independently estimate travel time between hubs and dwell time within hubs for type II parcels. The model leverages historical data and real-time updates to improve prediction accuracy. The total travel time results from summing all travel and dwell times, enabling dynamic routing. By incorporating these elements, our approach aims to provide a more accurate and practical solution for forecasting parcel delivery times within a logistics network.

Type I parcels, which have not yet entered the logistics network at the time of observation, present unique challenges for forecasting due to the lack of comprehensive status information. Consequently, demand forecasting for these parcels primarily relies on historical data. Although the literature on parcel arrival forecasting within logistic hubs is expanding, there remain significant gaps, particularly in terms of data granularity and forecasting horizons.

Existing studies, such as those by \cite{buckley2021hyperconnected}, have pioneered the use of real-time data to update forecasts for parcels approaching their final destination, yet these often exclude unordered parcels. This selective focus contrasts with broader forecasting efforts in other domains like tourism, where researchers employ time-series models, including exponential smoothing and seasonal ARIMA, to predict trends based on historical data and internet search indexes (\cite{lim2001forecasting}, \cite{kulendran2005modeling}, \cite{sun2019forecasting}).

A recent study by \cite{el2022deep} has proposed a comprehensive framework for processing historical transaction data to determine optimal training window sizes using auto-correlation and partial auto-correlation functions. Their innovative approach employs both Convolutional Neural Networks (CNN) and Recurrent Neural Networks (RNN) to forecast inbound container volumes, utilizing univariate and multivariate time series analyses to optimize forecasting outcomes. While their innovative use of deep learning model demonstrates high accuracy, the focus on daily data aggregation limits its applicability to the high-frequency, fine-grained forecasting needs of logistic hubs.

Our research directly addresses these limitations by proposing a model that forecasts parcel arrivals at 15-minute intervals, significantly extending the forecasting horizon from \( t+1 \) to \( t+96 \). Unlike \cite{el2022deep}'s reliance on CNNs and RNNs, our study utilizes an Artificial Neural Network (ANN) model, tailored to the constraints of our data which does not support the complexity required for more elaborate deep learning models. Additionally, we innovate by integrating an ensemble model that synthesizes dynamic real-time information, enhancing forecast reliability beyond what is achievable with direct ANN or Holt-Winters forecasts.

\begin{figure}[h]
    \centering
    \begin{subfigure}[b]{0.45\linewidth}
        \centering
        \includegraphics[width=\linewidth]{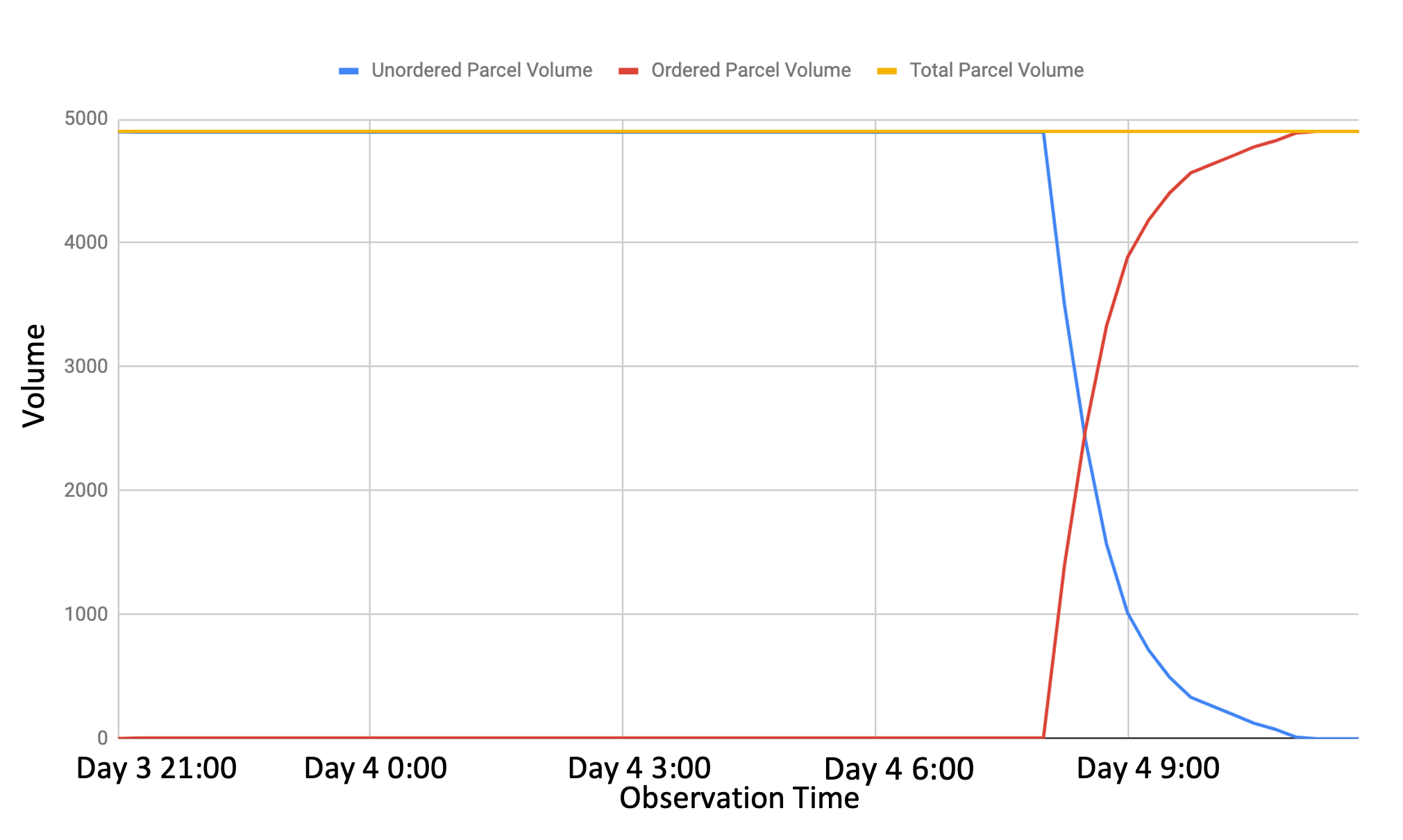}
        \caption[Actual Parcel Arrival Volumes]{}
        \label{fig:firstFig}
    \end{subfigure}
    \begin{subfigure}[b]{0.35\linewidth}
        \centering
        \includegraphics[width=\linewidth]{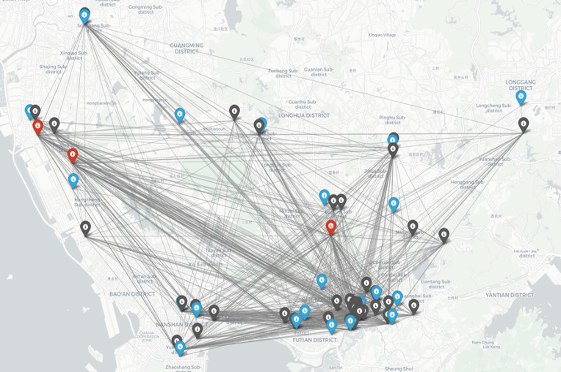}
        \caption[An Example of Parcel Logistic Hub]{}
        \label{fig:fig2}
    \end{subfigure}
    \caption{(a) The plot illustrates the actual parcel arrival volumes for a particular hub, with the same arrival time but at different observation times. The count of the unordered parcels is represented by the blue line. The amount of parcels that were ordered  is shown by the red line. Last but not least, the yellow line illustrates the total volume of the parcel. (b) An example of parcel logistic hub network in China. The grey dots represents the access hubs, blue dots are the local hubs and the red one is the gateway hubs. Grey lines are the paths between hubs.}
\end{figure}

Moreover, this approach is particularly suited to the intricacies of unordered parcel volumes, which do not conform to traditional time-series forecasting models due to their irregular arrival patterns. Figure \ref{fig:firstFig} illustrates the fluctuating volumes of unordered parcels as the arrival time approaches, underscoring the need for a forecasting approach that can adapt to such variability.

Finally, our research introduces a sequence-to-sequence neural network model (ANN) optimized for multi-horizon forecasting of unordered parcel volumes. This model, complemented by an ensemble method, effectively combines forecasts of ordered and unordered parcels to predict total arrival volumes, setting a new benchmark for accuracy in logistic hub operations. This method not only aligns with findings from neural network applications in other forecasting domains (\cite{kline2004methods}, \cite{taieb2015bias}, \cite{nor2018hybrid}, \cite{wang2018modelling}) but also significantly advances the capability to manage high-frequency, high-variability demand scenarios typical of modern logistic environments.

\section{Methodology}
In this section, we approach our parcel arrival forecast problem as a combination of unordered parcel arrivals forecast and ordered parcel arrival predictions.  First, we describe the background of the method, including the hubs in the logistic network, parcel definitions, and updates for the dynamically updated arrival forecasts. Then, we cover the arrival volume prediction for unordered and ordered parcels. In Section \ref{sec: ensemble}, we describe how to combine the arrival forecasts from the two types of parcels. Finally, we describe the algorithm for probability forecasting of parcel information in Section \ref{sec: parcel info}.

\subsection{Overview} \label{sec: background}

Hubs and vehicle-based routes are essential elements of a typical parcel logistics network, as described by \cite{buckley2021hyperconnected}. Such a network is denoted by \( N(H,R) \), where \( h \in H \) represents the set of hubs and \( r \in R \) indicates the set of routes. Figure \ref{fig:fig2} depicts a logistics network modeled based on one of China's Megacities, showcasing three distinct types of hubs: access hubs, local hubs, and gateway hubs, as categorized by \cite{montreuil2018urban}. Each type of hub plays a critical role within the network. Moreover, any of these hubs can act as a target hub, where parcels either enter the network or are transferred en route to their final destination. This adaptability makes our forecasting methods applicable to any hub or path within the logistics network.

The objective of this work is to provide forecasts of the future workload $\hat{c_t} \in \hat{C}$ for each time period $t$ with length of interval $I$, where $\hat{c_t}, t = 0,..., T$, represents the quantity of parcel arrivals during period $t$ and $T$ is the forecast horizon. Here, $t$ is specified relative to the observation time, which means that every $I$, period $t$ will refer to the forecast period that is $[tI, (t+1)I)$ into the future. For example, if we set $I = 15$ minutes and $T = 95$, then $\hat{c_t} \in \hat{C}$ represents the set of arrival forecasts for the next 24 hours at 15-minute intervals.

We categorize parcels into two types based on their status at the observation time for the target hub: Type I parcels (unordered) and Type II parcels (ordered but not yet arrived at the hub \(h \in H\)). Type I parcels are those that have not been ordered yet but are expected to arrive at hub \(h \in H\) within the forecast period \( t \), ranging from \( t = 0 \) to \( T \). The parcel types for a given parcel are not predetermined. As time progresses, parcels initially classified as Type I are ordered and transition to Type II, where their route and origin-destination information become known. This dynamic categorization allows for precise adjustments to our forecasting as parcel statuses evolve from unordered to ordered within the logistics framework.

\begin{figure}
    \centering
	\includegraphics[width=0.8\textwidth]{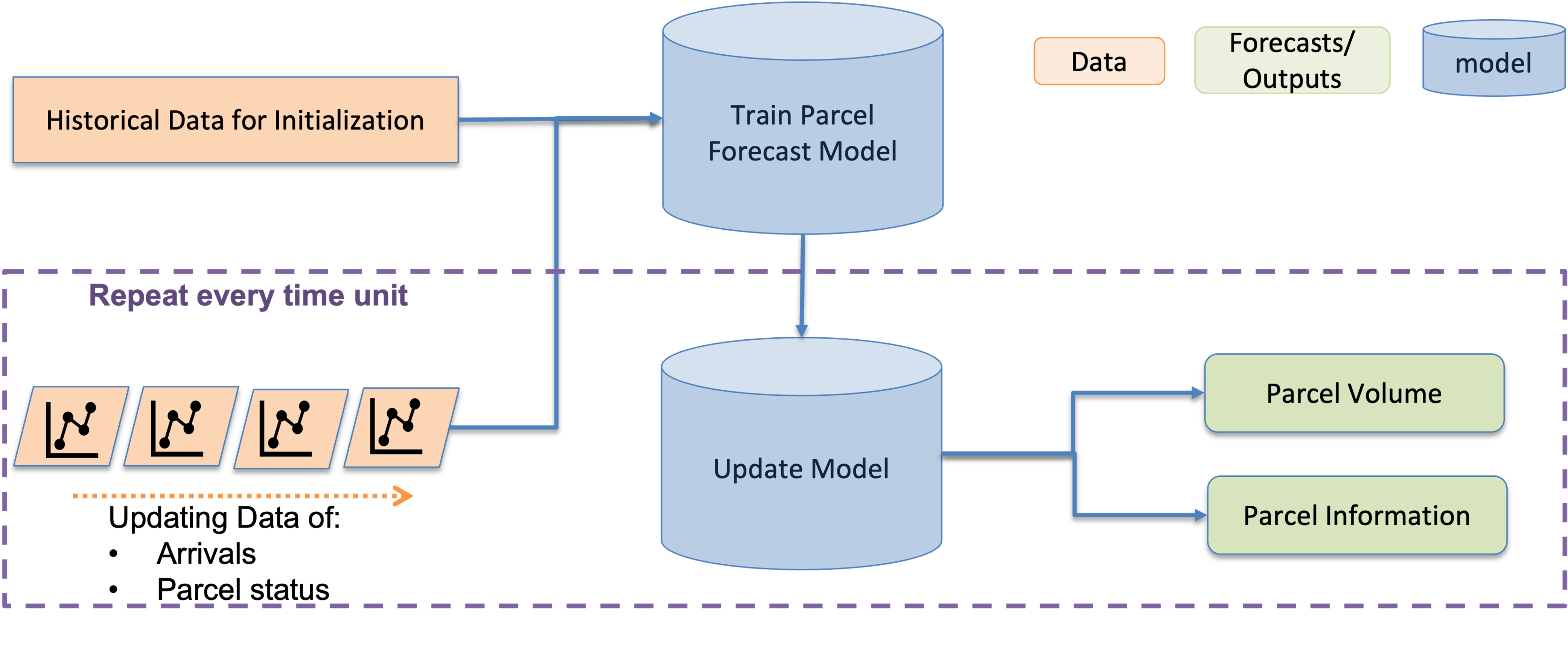}
	\caption[Dynamic Updating Model Framework]{Dynamic Updating Model Framework}
	\label{fig:fig3}
\end{figure}

Our forecasting approach incorporates dynamic updating to enhance prediction accuracy as real-time data becomes available. This mechanism allows the forecasting intervals to be adjusted every \(I\) minutes, denoted as \(t = 0, ..., T\), to reflect the forthcoming intervals \([tI, (t+1)I)\). Figure \ref{fig:fig3} illustrates a high-level concept of our dynamic updating framework. Initially, the model is configured with historical data. Subsequently, at each \(I\)-minute interval, it integrates new data on parcel arrival volumes, travel times, and status, updating the forecasts accordingly.

For example, consider the case of a Type II parcel with a known path \(h_i \rightarrow h_j \rightarrow h_k\), where \(h_k\) is the target hub, and the parcel is currently at \(h_i\). The predicted arrival time \(\hat{a}_p\) at the target hub is computed by summing the estimated travel times and dwell times at intermediate hubs:
\begin{equation}
\hat{a}_p = t_o + \hat{t}_{i,j} + \hat{t}_{j,k} + \hat{t}_{i} + \hat{t}_{j},
\end{equation}
where \(t_o\) is the observation time, \(\hat{t}_{i,j}\) represents the estimated travel time from hub \(h_i\) to \(h_j\), and \(\hat{t}_{i}\) is the estimated dwell time at hub \(h_i\).

As new data becomes available, the model is refreshed to reflect updated travel and dwell times. For example, once the parcel reaches \(h_j\), the new estimated arrival time at \(h_k\) is updated to:
\begin{equation}
\hat{a}_p = t_o' + \hat{t}_{j,k} + \hat{t}_{j},
\end{equation}
where \(t_o'\) is the new observation time, and \(t_o' > t_o\). This update utilizes the latest forecasts for travel time between \(h_j\) and \(h_k\) and the updated dwell time at \(h_j\).

This dynamic updating process ensures that our model remains responsive to the latest available data, providing forecasts that adapt to new developments within the logistics network every \(I\) minutes.

A similar updating process occurs for Type I parcels. At each observation time \(t_o\), the model uses newly available data to update the forecasts. For instance, the model will update based on the latest information on arrivals, effectively forecasting the parcel volumes that have not yet been processed or entered the network. This continual updating supports precise forecasting, essential for handling the dynamic and often unpredictable flows characteristic of logistics hubs.

In this research, we tackle the parcel arrival forecast problem using a novel multi-model approach, wherein the problem is decomposed into smaller, manageable sub-models that are independently trained and validated, then integrated into an ensemble network model. This methodology is articulated through two primary components of our framework: the forecast of unordered parcels (arrival volume forecast, detailed in section \ref{sec: unordered}) and the forecast of ordered parcels (arrival time prediction, discussed in  section \ref{sec: ordered}).

This model decomposition allows each sub-model to specialize in different aspects of the data and to utilize distinct sets of features effectively. For instance, we differentiate between types of parcels to maximize the use of data characteristics unique to each type. Furthermore, this approach facilitates capturing the correlations between the sub-models in the ensemble model, enhancing the overall forecasting performance.

Ensemble learning, or the multi-model ensemble combination technique, which involves breaking down complex models into simpler sub-models and then combining them, has been successfully applied in fields such as climate research and traffic modeling as evidenced by \cite{strobach2015improvement, petersen2019multi}. However, to the best of our knowledge, this method has not yet been explored in the context of parcel demand forecasting within logistics hubs.

In section \ref{sec: unordered}, we elaborate on our methodology for predicting the arrival volume of unordered parcels at the observation time $t_o$. Subsequently, in section \ref{sec: ordered}, we describe how we forecast the volume of ordered parcels. By synthesizing these forecasts, we are able to estimate the total volume of parcels arriving at the target hub:
\begin{equation}
\hat{c}_t = e(\hat{u}_t + \hat{o}_t), \quad t = 0, \ldots, T
\end{equation}
where $\hat{u}_t$ and $\hat{o}_t$ denote the forecasts for unordered and ordered parcels, respectively, relative to the same observation time. The function $e(*)$, representing our ensemble method, is detailed in section \ref{sec: ensemble}. This dynamic integration of forecasts from different types of parcels not only leverages the strengths of each model but also aligns with the dynamic nature of logistics operations, ensuring that our predictions remain relevant and timely.

\subsection{Arrival Volume Forecast - Type I Unordered Parcels}\label{sec: unordered}
In this section, we delineate our methodology for predicting the arrival volume of Type I, unordered parcels. Our aim is to estimate the number of parcels that have not been ordered as of the given observation time \( t_o \), but are anticipated to arrive at the target hub during specific future time intervals. These intervals are denoted as \( t \), where \( t = 0, \ldots, T \). Each interval \( t \) spans a predefined length \( I \) minutes, and is dynamically updated every \( I \) minutes to represent the subsequent interval \([tI, (t+1)I)\). This approach frames our forecasting task as a multi-step ahead challenge, necessitating precise and dynamic predictions based on evolving data inputs.

Given the limitations of traditional time series models for our context, as discussed in Section 1.2, and the propensity for error accumulation in step-by-step machine learning models, we opt to utilize an Artificial Neural Network (ANN) model for this task. The ANN model is designed to forecast the unordered parcel arrival volume \( \hat{U} = [\hat{u}_0, \hat{u}_1, \ldots, \hat{u}_T] \) as a function of:
\begin{equation}
\hat{U} = f(U, x^h, x^s)
\end{equation}
where \( U \) denotes the historical data of unordered parcel arrivals at the selected hub, \( x^h \) represents the historical temporal covariates (e.g., total historical arrival volumes), and \( x^s \) includes static, time-invariant features. The function \( f \) is determined by the neural network architecture.

The ANN model's capability to generate multi-step forecasts and its proficiency in capturing non-linear relationships within the data make it particularly suited for this forecasting problem. Unlike traditional approaches requiring detailed feature engineering, ANNs can approximate a wide range of functions accurately, facilitating the handling of complex patterns without extensive pre-assumptions. As noted in \cite{zhang1998forecasting, zhang2003time}, ANNs excel in pattern recognition and classification, making them ideal for time series modeling and forecasting applications. Although the unordered parcel data does not conform to typical time-series data, it exhibits seasonal patterns relative to each observation point, justifying the use of ANNs in this scenario. 

 In this paper, we have demonstrated the use of ANN models for predicting the arrival of unordered parcels. However, there are various options for neural network models that can be used for this purpose. Considering the seasonal nature of the data and its observation time, NN models capable of handling time-series data can be employed as well. A one-dimensional convolutional neural network (CNN) that can recognize complex data patterns as shown in \cite{zhao2017convolutional} is one such model that may be suitable. Moreover, recurrent neural network (RNN), which is a common and widely used architecture for time series forecasting models, and its extensions, such as long short-term memory (LSTM) and gated recurrent unit (GRU) models, could also be used. Furthermore, encoder-decoder and attention mechanisms can be applied in this regard to enhance the prediction accuracy.

Consider an ANN model with two hidden layers, which allows for the extraction and integration of more complex data patterns. This architecture is particularly effective for datasets that exhibit intricate seasonal and non-linear dynamics, such as the arrival times of unordered parcels. The relationship between each of the outputs $\hat{u}_t \in \hat{U}$ and the set of inputs $X = \{U, x^h, x^s\}$ can be expressed through the following model equation:
\begin{equation} \label{equation: ann_two_layer}
    \hat{u}_t = \alpha_0 + \sum_{j=1}^{n_h^1} \alpha_j g\left(\beta_{0j} + \sum_{k=1}^{n_h^2} \gamma_{jk} g\left(\delta_{0k} + \sum_{i=1}^{n_i} \delta_{ik} x_i\right)\right), \quad t = 0, \ldots, T
\end{equation}
where:
- $\alpha$, $\beta$, $\gamma$, and $\delta$ are the parameters of the model, representing the weights between the input, hidden, and output layers.
- $n_h^1$ and $n_h^2$ are the number of neurons in the first and second hidden layers, respectively.
- $g(*)$ is the activation function in both hidden layers, typically the rectified linear activation function (ReLU) for its properties in mitigating the vanishing gradient problem:
\[
g(z) = \max(0, z)
\]

This two-layer structure enhances the neural network’s capability to perform abstract pattern recognition, allowing it to learn more complex relationships within the data. The double layer of non-linear transformation equips the model to better understand and predict patterns that are difficult to capture with simpler models.

\begin{figure}
    \centering
	\includegraphics[width=0.5\textwidth]{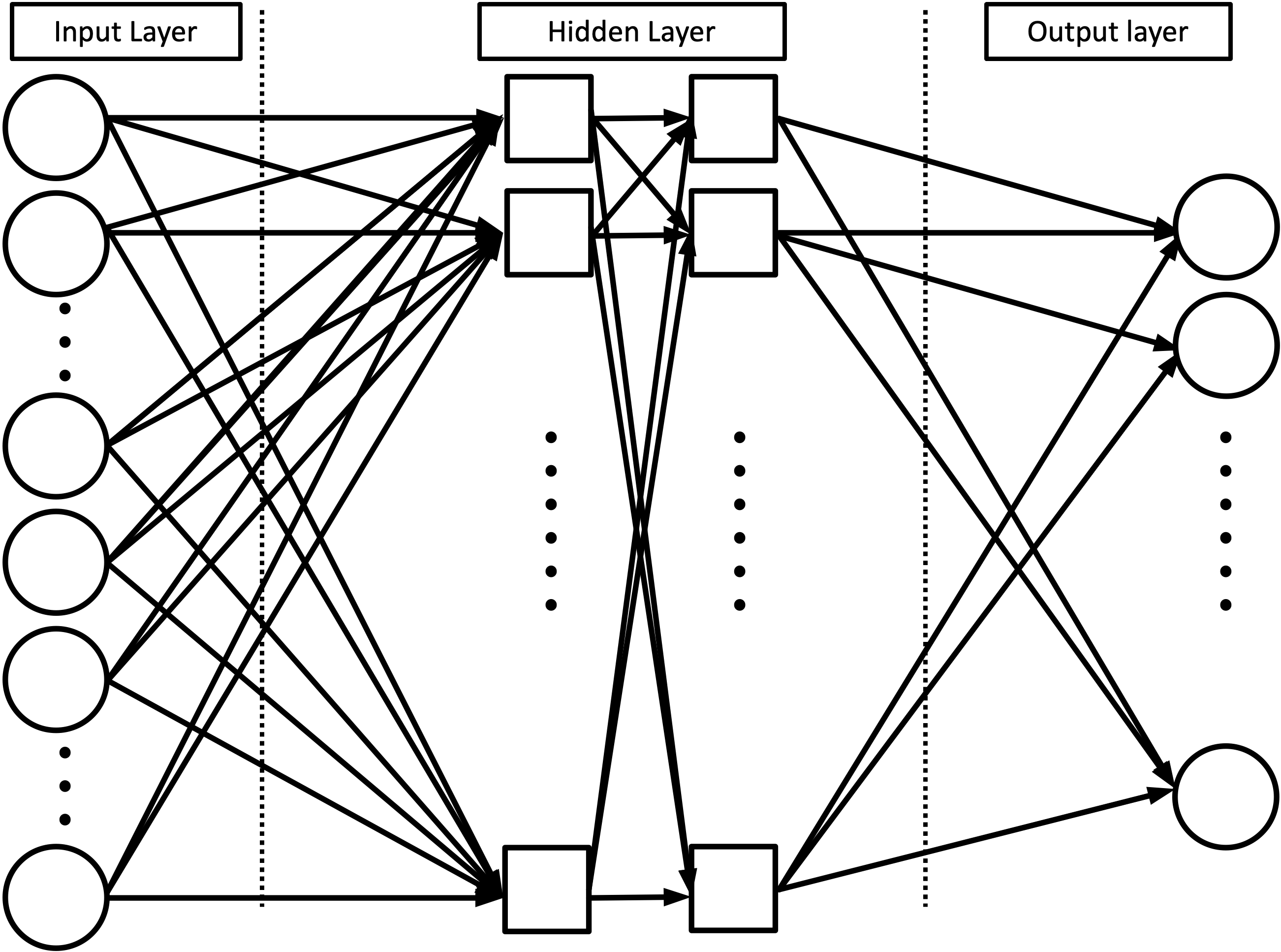}
	\caption[ANN model Architecture]{The plot shows the architecture of a feed-forward network with 2 hidden layers}
	\label{fig:fig4}
\end{figure}

Figure \ref{fig:fig4} provides an illustrative example of the ANN model architecture, showcasing how each layer connects and contributes to the final prediction. Given the complex nature of the data and its inherent variability, employing a two-layer ANN model enhances the forecast's reliability by capturing deep-rooted patterns in the data, making it particularly suitable for the high-frequency, multi-step forecasting required in logistic hub operations.

\subsection{Arrival Volume Prediction - Type II Ordered Parcels}\label{sec: ordered}
This section delineates our methodology for forecasting the arrival volume of Type II parcels, which are already within the logistics network but have not yet reached their target hub at the observation time. Unlike Type I parcels, Type II parcels offer the advantage of available sequential arrival data at previous hubs and known planned paths for future arrivals. This accessible information facilitates a more precise prediction of their arrival volumes, which in turn aids logistics operators in optimizing resource allocation and operational planning.

The forecasting approach for Type II parcels integrates two crucial components: arrival time prediction and arrival volume prediction. Accurately estimating the arrival time for each Type II parcel is essential, as it directly influences the computation of the arrival volume \( o_t \) during specific periods \( t \). The challenge primarily lies in accurately predicting each parcel's travel time, which, when summed with the current observation time, provides the arrival time.

\begin{equation}
\hat{a}_p = t_o + \sum \hat{t}_{travel} + \sum \hat{t}_{dwell},
\end{equation}

where \( \hat{a}_p \) is the predicted arrival time at the target hub, \( t_o \) is the observation time, \( \hat{t}_{travel} \) are the predicted travel times for each subsequent hub-hub segment, and \( \hat{t}_{dwell} \) are the predicted dwell times at each intermediate hub. The model thus aggregates dynamically updated travel and dwell time predictions, utilizing the parcels' current location and planned path data.

Our proposed model operates on a dynamic updating principle, where the model is retrained and its predictions updated every \( I \) minutes to integrate new data. For Type II parcels, this includes adjustments based on changes in parcel location and transit status. This enables continuous refinement of the arrival time and volume predictions:

\begin{equation}
\hat{o}_t = \text{Count of parcels predicted to arrive in interval } [tI, (t+1)I).
\end{equation}

As new information becomes available, such as changes in parcel status or updates in transit conditions, the model adapts, recalculating the predicted arrival times and volumes to maintain forecasting accuracy. This iterative updating mechanism ensures that predictions remain relevant and are reflective of the latest available data, supporting more effective decision-making in the management of logistics operations.

Random Forest, a robust algorithm developed in \cite{breiman2001random}, leverages an ensemble of decision trees to predict outcomes for both classification and regression problems effectively. The algorithm operates by constructing a multitude of decision trees during training and outputs the average prediction of the individual trees for regression tasks, which enhances prediction accuracy and controls over-fitting as shown in \cite{biau2016random}. 

\begin{figure}[ht]
    \centering
	\includegraphics[width=0.6\textwidth]{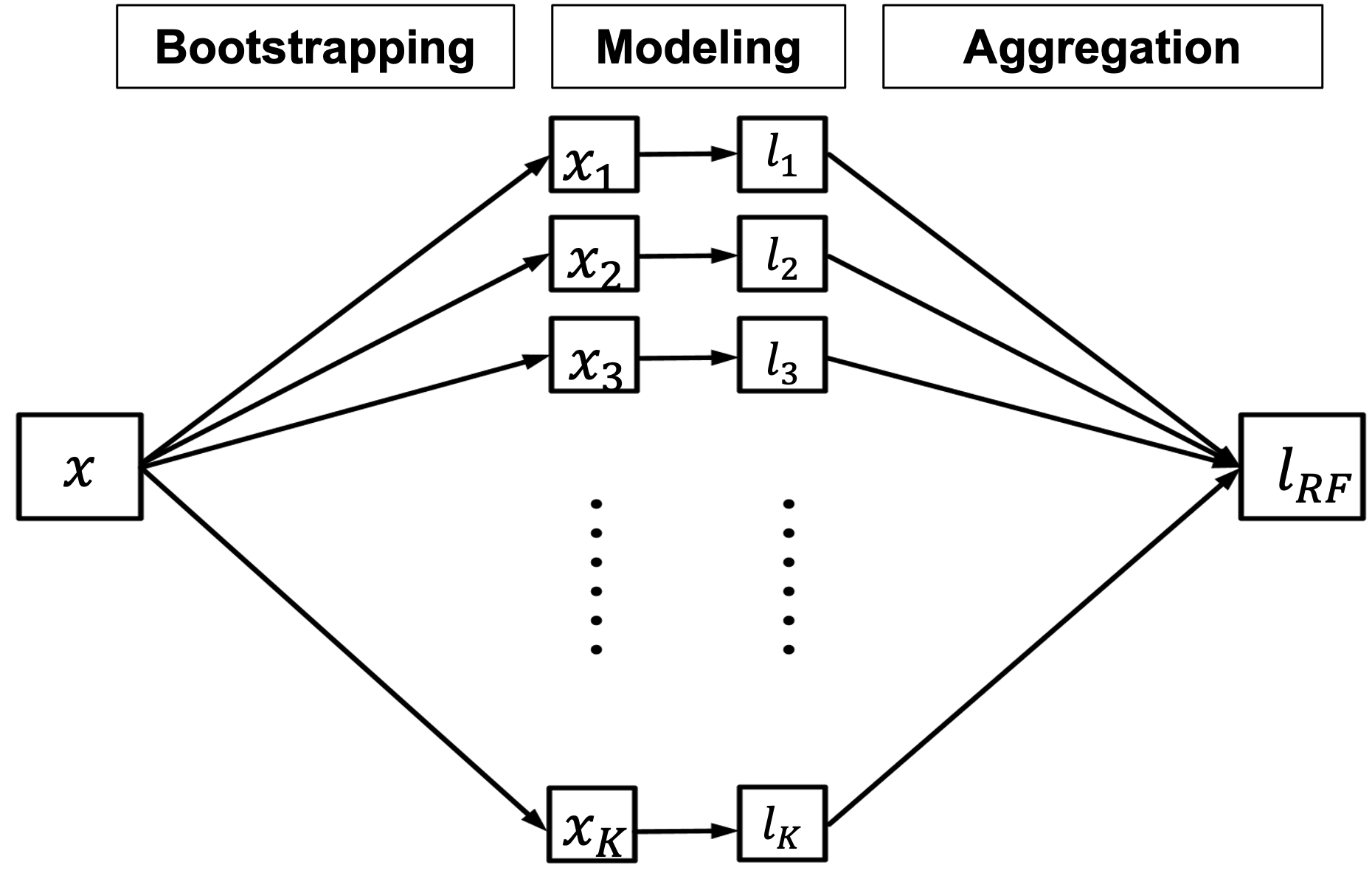}
	\caption[Random Forest Model with $K$ Decision Trees]{The plot shows the architecture of a random forest model, where $x$ represents input instance from the historical data used for forecasting the travel/dwell time. $x_k$ represents $k$-th sample of the data, $l_k$ is the $k$-th decision tree and $l_{RF}(*)$ is the final predicted travel/dwell time.}
	\label{fig:fig6}
\end{figure}

The architecture of a random forest model is illustrated in Figure \ref{fig:fig6}.  Given a feature space \(\mathcal{X}\) and an output space \(\mathcal{Y}\), Random Forest seeks to learn a mapping based on a training set \(\mathcal{D}\) of \(n\) instances, each with \(d\) features. It builds \(K\) decision trees where each tree \(k\) provides a prediction \(l_k(x)\) for an input instance \(x\). The ensemble prediction, denoted by \(l_{RF}(x)\), is calculated as:

\begin{equation}
l_{RF}(x) = \frac{1}{K} \sum_{k=1}^{K} l_k(x),
\end{equation}

where \(l_{RF}(x)\) represents the estimated travel or dwell time for our context. The construction of each tree involves selecting a random subset of \(m\) features (\(m < d\)) to ensure diversity among the trees, thus enhancing the model's robustness and accuracy. The training process utilizes measures such as Gini impurity or information gain to optimally split the data at each node of the trees.

In the context of our research, the Random Forest model is employed to estimate the travel and dwell times for Type II parcels. It uses a range of features including the time of day, day of the week, and specific location IDs of the hubs. These predictions inform the calculation of expected arrival times and volumes at subsequent hubs, thus aiding in resource allocation and operational planning. 

Future iterations of our model could incorporate additional features to further enhance the accuracy of the predictions. Depending on factors such as the retraining costs and expected improvements in forecast precision, the model may be retrained periodically every \(I\) with new data collected from the logistics network.
\subsection{Ensemble Model} \label{sec: ensemble}

\begin{figure}[h]
    \centering
    \includegraphics[width=0.7\textwidth]{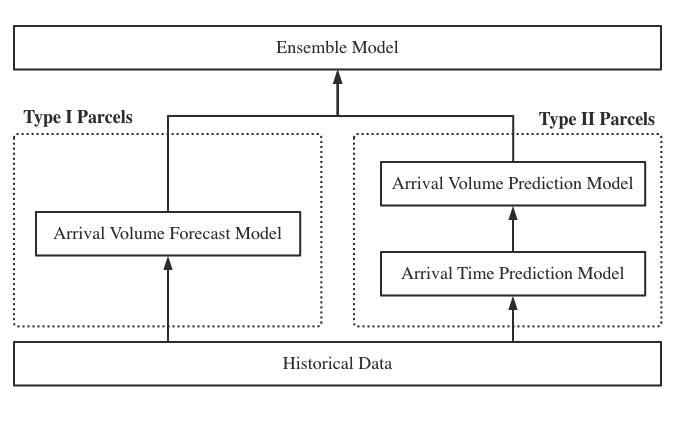}
    \caption[Ensemble Model]{The structure of the ensemble model combining forecasts from unordered and ordered parcel models.}
    \label{fig:fig5}
\end{figure}

Having described the methodologies for forecasting unordered and ordered parcel arrival volumes independently, we now address how to integrate these predictions to estimate the total arrival volume at the target hub. The framework for our ensemble model is illustrated in Figure \ref{fig:fig5}. This approach aligns with ensemble learning principles, where multiple sub-models with potentially diverse input data characteristics are combined to enhance predictive performance.

The simplest form of integrating the forecasts from the two sub-models (unordered and ordered parcels) is through direct summation. This is based on the premise that the total parcel arrival volume $\hat{c}_t$ at any time interval \(t\) is the sum of the predicted volumes of unordered and ordered parcels arriving during that interval:
\begin{equation}
\hat{c}_t = \hat{u}_t + \hat{o}_t, \quad t = 0, \ldots, T
\end{equation}
This approach assumes independence between the arrival volumes of unordered and ordered parcels and relies on the additive nature of their impacts.

While summation provides a straightforward method for combining forecasts, capturing the underlying correlations between unordered and ordered parcel arrivals can enhance prediction accuracy. To this end, we employ a fully connected neural network as our ensemble method, which allows for the integration of outputs from both sub-models more dynamically:
\begin{equation}
\hat{c}_t = \gamma_0 + \sum_{j=1}^{n_h'} \gamma_j g\left(\delta_{0j} + \sum_{i=1}^{n_i'} \delta_{ij} y_i\right), \quad t = 0, \ldots, T
\end{equation}
Here, \(Y = \{\hat{U}, \hat{O}\}\) represents the concatenated outputs from the unordered and ordered parcel models respectively, with \(y_i \in Y\) being the inputs to the neural network. The parameters \(\gamma\) and \(\delta\) are the network weights, \(n_h'\) denotes the number of neurons in the hidden layer, and \(n_i'\) the number of input nodes, with \(g(*)\) representing the ReLU activation function.

This neural network-based approach not only combines the forecasts but also captures potential interactions between the different types of parcel data, thus providing a more nuanced and potentially more accurate forecast of total parcel arrivals.

The choice between simple summation and neural network-based ensemble can be dictated by the specific requirements of accuracy and computational efficiency in operational settings, as well as the availability of data to train more complex models.

\subsection{Parcel Destination Share Forecast} \label{sec: parcel info}

The accuracy in forecasting the distribution of parcel destinations is crucial for logistical planning, especially for managing resources at hubs. This section presents our methodology for predicting the destination shares of unordered parcels, which is vital for sorting and dispatching processes. For example, if managers have the forecast about the number of parcels arriving at various destinations, they can properly manage staffing, particularly for the door-to-door parcel sorting process, which will bring huge benefits to the logistic hubs.

The methodology as shown in Algorithm \ref{alg:weighted_average_od_pair} comprises two stages: initialization and updating, delineated by the observation time $t_o$. Initially, we utilize historical data from earlier periods to estimate the share of parcels destined for each location. This is formalized as:
\begin{equation}
d_{t}^j = \frac{m_t^j}{m_t},
\end{equation}
where $d_{t}^j$ is the estimated destination share for destination hub $h_j$ during time period $t$ given the observation time $t_o$, $m_t^j$ is the count of unordered parcels for $h_j$, and $m_t$ is the total count of unordered parcels.

In the updating stage, new data available at each interval \(I\) is used to refine these estimates dynamically. The updated share $\rho_t^j$ is calculated from new incoming data, and the destination share $d_{t}^j$ is adjusted using a smoothing parameter $\alpha$, ensuring the estimates remain responsive to recent trends:
\begin{equation}
d_{t}^j = \alpha d_{t-1}^j + (1 - \alpha) \rho_t^j,
\end{equation}
with normalization enforced after each update to maintain probabilistic constraints:
\begin{equation}
\sum_{h_j \in H_J} d_{t}^j = 1, \forall t = \{0, ..., T\}.
\end{equation}
Using these destination shares and the forecasted total unordered parcel volume $\hat{u_t}$, we can estimate the volume of arrivals for each destination:
\begin{equation}
\hat{u}_t^j = \hat{u_t} d_{t}^j.
\end{equation}

This detailed approach to forecasting destination shares not only aids in operational planning but also enhances the accuracy of the overall forecasting model by integrating dynamic updates and fine-grained parcel distribution predictions.

\begin{algorithm}
\caption{Probability Forecast using Weighted Average - OD Pair}\label{alg:weighted_average_od_pair}
\begin{multicols}{2}
\begin{algorithmic}[1]
\REQUIRE 
$u_{t, t_o}^j$: the volume of unordered parcels arrivals at the target hub during time period $t$ at observation time $t_o$ with destination hub $h_j \in H_J$.\\
$J$: the set of destination hubs for the target hub.\\
$T_I$: the set of initialization observation times.\\
$T_U$: the set of updating observation times.\\
$\alpha$: smoothing parameter, where $0\leq \alpha \leq 1$.

\ENSURE The set of estimated destination share $d_{t}^j \in D, t\in \{0, ..., T\}$.

\STATE \COMMENT{Initialize}
\FOR{$t \in \{0, ..., T\}$}
    \STATE $m_t \gets 0$.
    \FOR{$h_j \in H_J$}
        \STATE $m^j_t \gets 0$.
        \FOR{$t_o \in T_I$}
            \STATE $m^j_t \gets m^j_t + u_{t, t_o}^j$.
        \ENDFOR
        \STATE $m_t \gets m_t + m^j_t$.
    \ENDFOR
    \FOR{$h_j \in H_J$}
        \STATE $d_{t}^{j} \gets m^j_t / m_t$.
    \ENDFOR
\ENDFOR

\STATE \COMMENT{Updating:}
\FOR{$t_o \in T_U$}
    \FOR{$t \in \{0, ..., T\}$}
        \STATE $m_t \gets 0$.
        \FOR{$h_j \in H_J$}
            \STATE $m_t \gets m_t + u_{t, t_o}^j$.
        \ENDFOR
        \FOR{$h_j \in H_J$}
            \STATE $\rho_t^j \gets u_{t, t_o}^j / m_t$.
        \ENDFOR
        \STATE $d^j_t \gets (1 - \alpha) \cdot d^j_t + \alpha \cdot \rho_t^j$.
        \STATE $d_{t} \gets 0$ \COMMENT{Normalization}.
        \FOR{$h_j \in H_J$}
            \STATE $d_{t} \gets d_{t} + d^j_t$.
        \ENDFOR
        \STATE $d^j_t \gets d^j_t / d_{t}$.
    \ENDFOR
\ENDFOR
\end{algorithmic}
\end{multicols}
\end{algorithm}

\section{Experiments}

In this paper, we evaluate the efficacy of our forecast updating model designed to predict parcel arrival volumes every 15 minutes for the next 24 hours. The aim is to generate forecasts for future workload \(\hat{c}_t\in \hat{C}\) at any given hub, for each time period \(t\) with a length of \(I = 15\) minutes, where \(t = 0, ..., T = 95\). Computational studies have been conducted to validate the effectiveness of our model, and this section details those investigations. We discuss the dataset used, the parcel logistics network simulator employed for the study, the forecasting techniques applied to both ordered and unordered parcels, and the alternative methodologies assessed for comparative performance analysis. Our objective is to demonstrate the enhanced accuracy of our model in improving parcel arrival forecasts within the logistics network.

\subsection{Data} 
For our experiments, we used a comprehensive thirty-day dataset detailing the complete travel history of parcels through the logistics network, facilitated by the Georgia Tech PI Lab Intracity Logistics Simulator, with details in \cite{kaboudvand2021}. This simulator integrates various components including customers, logistic hub decision-makers, movers, and packages, which interact within an urban parcel logistics framework. The data encompasses detailed information on all parcels processed during the study period, including real-time positions, origins and destinations, and pickup and arrival times at various hubs.

The first 27 days of the dataset served as the training and validation data for model training and parameter tuning. The effectiveness of our forecasting approach was then assessed using the data from the last 3 days. This phase focused on evaluating the performance of our forecasting model by comparing its error metrics with those of alternative models, aiming to underscore the superior accuracy of our method in predicting parcel arrival volumes. The use of separate datasets for training and testing helps verify the robustness of our approach and ensures its generalizability to new, unseen datasets.

\subsection{Unordered Parcel Forecasting}
This section discusses our methodology for forecasting the arrival of Type I, unordered parcels, denoted as \(\hat{c}_t, t=0, ..., T = 95\). Each time interval \(t\) spans a fixed duration of \(I = 15\) minutes. As highlighted in section \ref{sec: unordered}, the challenge with unordered parcels arises from the non-time-series nature of their arrival data, despite its seasonal behavior. This limitation necessitates a forecasting approach that transcends conventional time-series models.

To effectively predict multi-horizon arrivals, we employ an Artificial Neural Network (ANN) model, leveraging historical arrival data. The ANN is designed to generate forecasts for every 15-minute interval over a 24-hour period from the specified observation time.

\subsubsection{Model Inputs}
The model incorporates two primary types of input features:
\begin{enumerate}
    \item \textbf{Previous Day's Unordered Parcel Volume:} This feature represents the actual arrival volume of unordered parcels from the previous day, providing a direct link to our target forecasting periods. For example, if the current observation time \(t_o\) is the start of day 2, we use the unordered parcel volume from day 1 as an input to predict arrivals for day 2.
    \item \textbf{Recent Total Parcel Volume:} This feature captures the total parcel arrivals at the hub during the 4 hours leading up to the current observation time \(t_o\). It helps the model grasp short-term fluctuations in arrival volumes, enhancing the responsiveness of the forecasts.
\end{enumerate}

Both input features are critical for capturing the dynamics of parcel arrivals—while the former helps in identifying daily patterns, the latter adjusts for any immediate changes that could affect the prediction accuracy.

\subsubsection{ANN Architecture and Training}
The ANN model used in this experiment features two hidden layers:
\begin{itemize}
    \item The first hidden layer contains 1024 neurons.
    \item The second hidden layer contains 512 neurons.
\end{itemize}
This architecture is optimized using the Adam optimizer with a learning rate of 0.05 and trained over 500 epochs to ensure thorough learning. To mitigate overfitting, a regularization technique involving a weight decay rate of 0.01 is applied.

The dynamic nature of our model allows for the continuous updating of input data and predictions, maintaining the relevance and accuracy of the forecasts as new data becomes available. By employing this robust ANN framework, we aim to demonstrate superior predictive performance in capturing the intricate arrival patterns of unordered parcels at logistic hubs.

\subsection{Ordered Parcel Forecasting}
\label{sec: ordered}
In this section, we discuss our approach to dynamically forecasting the arrival volume of Type II, ordered parcels at the target hub. This involves accurate predictions of dwell times at hubs and travel times on routes, which are crucial for estimating when parcels will arrive at their destinations. Such forecasts are vital for efficient resource allocation and planning at logistic hubs.

Initially, we identified all feasible paths that a parcel might take within the logistics network, including all nodes (hubs) and links (routes) as shown in Table \ref{tbl:inputForRF}. This subset was derived from the historical data of parcels' travel routes. We focused on predicting two main components:
\begin{enumerate}
    \item \textbf{Dwell Time at Hubs:} Time a parcel is expected to spend at each hub after its arrival.
    \item \textbf{Travel Time on Routes:} Time taken for a parcel to travel between consecutive hubs on its route.
\end{enumerate}

Historical data on dwell times at hubs and travel times on routes were collected. Using this data, we applied a Random Forest model configured to dynamically forecast these times for future periods. The model was trained with 100 estimators on subsets of the data, using features such as location ID, time of day, and day of the week, among others.

\begin{table}[h]
\centering
\caption[Input Data for Random Forest Model]{Example of Input Data for Random Forest Model}
\label{tbl:inputForRF}
\resizebox{\textwidth}{!}{%
\begin{tabular}{lllllll}
\hline
Nodes or links  & Arrival time & Departure time & Time of day & Day of week & Dwell time & Travelling time \\ \hline
GH1 & 4:21AM & - & 4 & Monday & 105.23 & -\\
AH5 & 12:26PM & - & 12 & Wednesday & 23.54 & -\\
AH12-LH4 & - & 2:56PM & 14 & Sunday & - & 84.75\\
LH8-GH1 & - & 8:12PM & 8 & Tuesday & - & 26.71\\\hline
\end{tabular}%
}
\end{table}

Using the forecasted dwell and travel times, we update the arrival time predictions for each parcel. This update occurs every 15 minutes, reflecting real-time changes in the parcel’s status within the logistics network. Whenever a parcel arrives at a hub or undergoes a status update, its predicted arrival time at the destination hub is recalculated. This calculation includes the latest forecasted dwell and travel times for its remaining path.

This dynamic prediction process is critical for maintaining the accuracy of the arrival forecasts, ensuring that they adapt to new developments in the logistics network as they occur. The continuous updating mechanism is outlined in Algorithm \ref{alg:cap}, providing detailed steps on how the predictions are adjusted in response to real-time data.

The combination of accurate travel and dwell time predictions with real-time updates enables our model to effectively forecast the arrival volumes of ordered parcels, thus enhancing the operational efficiency of logistics hubs.

\begin{algorithm}
\caption{Dynamic Parcel Arrival Prediction for Type II Parcels}\label{alg:cap}
\begin{algorithmic}[1]
\REQUIRE Historical arrival data, the travelling path of type II parcels in the system.
\ENSURE The predicting numbers of parcels arrived at target hub.

\FOR{$t$ within planning horizon}
    \STATE Predict travel time and dwell time given $t$ and historical data.
    \FOR{all type II parcels in the system}
        \STATE Predict parcel's arrival time given its route and time prediction.
    \ENDFOR
    \STATE Update the predicted number of parcels arrived at the target hub every $I$ minutes.
    \STATE $t \gets t + 1$.
\ENDFOR
\end{algorithmic}
\end{algorithm}

\subsection{Methodologies for Comparison}
\label{sec: comparison_methods}
This section describes the alternative forecasting methods we employed as benchmarks to evaluate the performance of our innovative ensemble model. Our model focuses on predicting the total arrival volumes of parcels, a task best suited for time-series analysis. We used both traditional statistical models and machine learning approaches for comprehensive comparison.

\subsubsection{Holt-Winters Method (Method 1)}
The Holt-Winters method, also known as triple exponential smoothing, is a robust forecasting technique widely used for time-series data that exhibits both trend and seasonal variations. It extends simple exponential smoothing to accommodate data with trend and seasonality. The model is defined by the following set of equations:

    
    
\begin{align}
    \text{Level equation:} \quad & \ell_t = \alpha (y_t - s_{t-m}) + (1-\alpha)(\ell_{t-1} + b_{t-1}), \\
    \text{Trend equation:} \quad & b_t = \beta (\ell_t - \ell_{t-1}) + (1-\beta) b_{t-1}, \\
    \text{Seasonality equation:} \quad & s_t = \gamma (y_t - \ell_t) + (1-\gamma) s_{t-m}.
\end{align}

where $\alpha$, $\beta$, and $\gamma$ are smoothing parameters, $y_t$ represents the observed value at time $t$, $\ell_t$, $b_t$, and $s_t$ are the estimated level, trend, and seasonal component at time $t$, respectively, and $m$ is the seasonal period.

\subsubsection{Artificial Neural Network (Method 2)}
As a second method, we implemented an Artificial Neural Network (ANN) model to directly forecast the total parcel volumes. Unlike our ensemble approach, this model does not differentiate between ordered and unordered parcels but treats all parcel data uniformly. This comparison aims to determine the effectiveness of segmenting parcel data according to their status (ordered vs. unordered) and the corresponding forecasting accuracy.

\subsubsection{Ensemble Model (Method 3)}
Our primary model, referred to as Method 3, integrates forecasts of ordered and unordered parcels using an ensemble approach. The model operates as follows:
\begin{itemize}
    \item An ANN model forecasts the volume of unordered parcels using historical data.
    \item A Random Forest model predicts travel times and converts these into volume forecasts for ordered parcels based on updated status information.
    \item The outputs from both models are combined using a summarization technique to produce the final forecast for the total parcel arrival volume at the target hub.
\end{itemize}

\subsubsection{Advanced Ensemble Model (Method 4)}
Method 4 is an updated version compared with method 3 that employs another ANN model instead of simple summarization. This ANN model, which consists of a single layer with 256 neurons, takes the forecasted arrival volume for unordered parcels and the predicted arrival volume for ordered parcels as input. The output is the expected arrival volume for total parcels. The model is trained using a learning rate of 0.01 to optimize performance. In addition to the parcel volume forecasts, we also incorporate calendar factors, such as the period of the day, hour of the day, and day of the week. By doing so, we aim to capture the correlation between ordered and unordered parcel arrival volumes and to account for the impact of calendar features, which could potentially improve our forecast accuracy.

By comparing these methods, we aim to validate the effectiveness of our ensemble model and demonstrate its superior performance in capturing complex dynamics and providing accurate parcel arrival forecasts.

\section{Computational Result}
To evaluate the effectiveness of our proposed model, we utilized the Mean Absolute Scaled Error (MASE) as a key performance indicator (KPI). The MASE is particularly suitable for time series analysis as it measures forecast accuracy relative to a naive benchmark, accommodating both the magnitude and direction of errors. This scale-independent measure allows comparisons across different data scales and is formulated as follows:

\begin{equation}
\text{MASE} = \frac{\text{Mean Absolute Error of the Forecast}}{\text{Mean Absolute Error of the Naive Forecast}}
\end{equation}

Where the Mean Absolute Error of the Forecast is calculated by:

\begin{equation}
\text{MAE} = \frac{1}{n} \sum_{i=1}^{n} |y_i - \hat{y}_i|
\end{equation}

And the Mean Absolute Error of the Naive Forecast is the average absolute difference between the actual values and their corresponding values in the previous period (naive forecast). Here, we utilize the previous day's same-time-period data as the naive forecast. Variables are defined as follows:

\begin{itemize}
    \item \( n \): the number of observations.
    \item \( y_i \): the actual value at the \( i \)-th observation.
    \item \( \hat{y}_i \): the forecasted value at the \( i \)-th observation.
\end{itemize}

A MASE value less than 1 indicates that the forecast model outperforms the naive benchmark, providing forecasts that are more accurate than those obtained by simply using the previous period’s data. Conversely, values greater than 1 suggest inferior performance compared to the naive forecast. MASE is a critical metric for comparing the accuracy of different forecasting models, with lower values denoting superior forecasting performance.

The current research evaluates the efficacy of various forecasting models, including Holt-Winters (Method 1), ANN-direct (Method 2), Ensemble with summation (Method 3), and Ensemble with ANN (Method 4), in predicting the demand for parcels within a logistics network. According to the experimental results presented in Table \ref{tbl:result1}, all models surpass the Naive model in forecasting accuracy. Notably, a comparative analysis between Method 1 and Method 2 underscores that non-linear models are more adept at handling the complex dynamics inherent in the data, thereby providing a robust and precise forecasting tool. Furthermore, comparisons between Method 2 and Method 3 illustrate that the integration of parcel status as an external factor, alongside distinctions between ordered and unordered parcels, slightly enhances forecasting accuracy. Crucially, the results demonstrate that the ANN-based ensemble method (Method 4) outperforms other models, delivering the highest accuracy in predicting parcel demand with the lowest forecasting errors. The utilization of an ANN model in an ensemble configuration effectively captures non-linear relationships and leverages the strengths of diverse models to improve predictive performance.

\begin{figure} [h]
    \centering
	\includegraphics[width=0.7\textwidth]{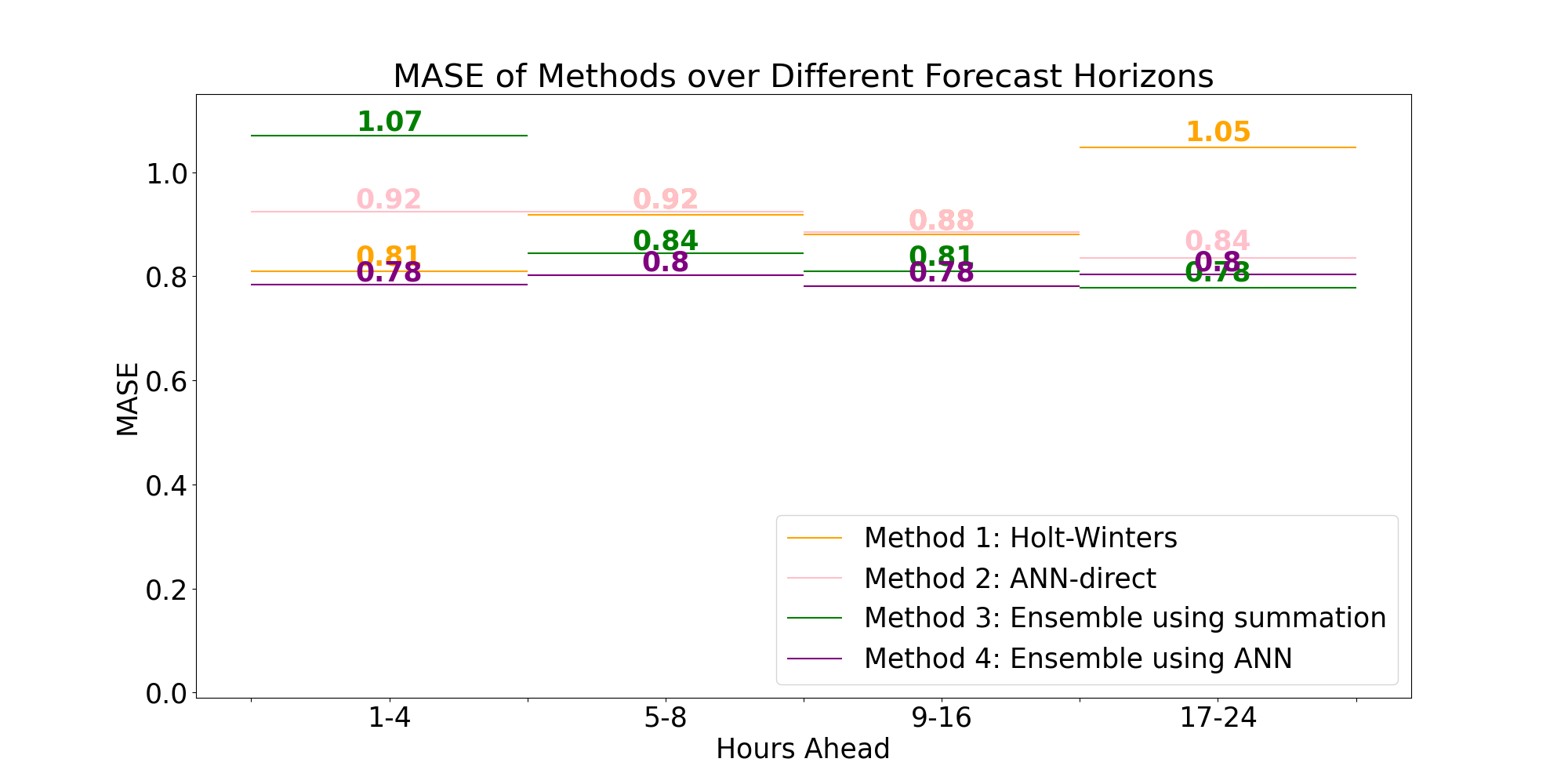}
	\caption[Horzion Results]{MASE of Methods over Different Forecast Horizons}
	\label{fig:fig8}
\end{figure}

Employing MASE as the error metric allows us to evaluate the general performance of various models. However, we are also keen on analyzing the performance of models at distinct time horizons. To achieve this, we partitioned our 96-period (24-hour) horizon into four categories: 1-4 hours ahead, 5-8 hours ahead, 9-16 hours ahead, and 17-24 hours ahead. We have determined that this division aligns with the four-hour workload shifts observed in the warehouse. The results of the evaluation of the Mean Absolute Scaled Error (MASE) for different time horizons and methods are presented in Figure \ref{fig:fig8}. Method 4 demonstrated superior performance across all horizons, indicating the robustness of our proposed approach. Notably, the Holt-Winters method exhibited poor performance when evaluated using the general MASE metric. However, it achieved relatively good accuracy for the horizon of 1-4 hours ahead, but the worst performance for the longest horizon, i.e., 17-24 hours ahead. Conversely, Method 3, which utilizes an ensemble approach through summation, demonstrated poor performance for the first horizon group but achieved the highest accuracy for the last horizon group. Interestingly, the forecast accuracy of Method 2 and Method 3 appeared to increase with the horizon length, which warrants further investigation. These findings highlight the potential benefits of our proposed method in achieving accurate forecasts across varying time horizons.

\begin{table}[width=.6\linewidth,cols=4,pos=h]
\caption{Comparison of model performance for dynamic parcel volume forecasting}\label{tbl:result1}
\begin{tabular*}{\tblwidth}{@{}LL@{} }
\toprule
Method & MASE \\ \midrule
Method 1: Holt-Winters & 0.92 \\
Method 2: ANN-direct & 0.88 \\
Method 3: Ensemble using summation & 0.86 \\
Method 4: Ensemble using ANN & \textbf{0.79} \\
\bottomrule
\end{tabular*}
\end{table}

\begin{figure} [h]
    \centering
    \makebox[\textwidth][c]{\includegraphics[width=1.2\textwidth]{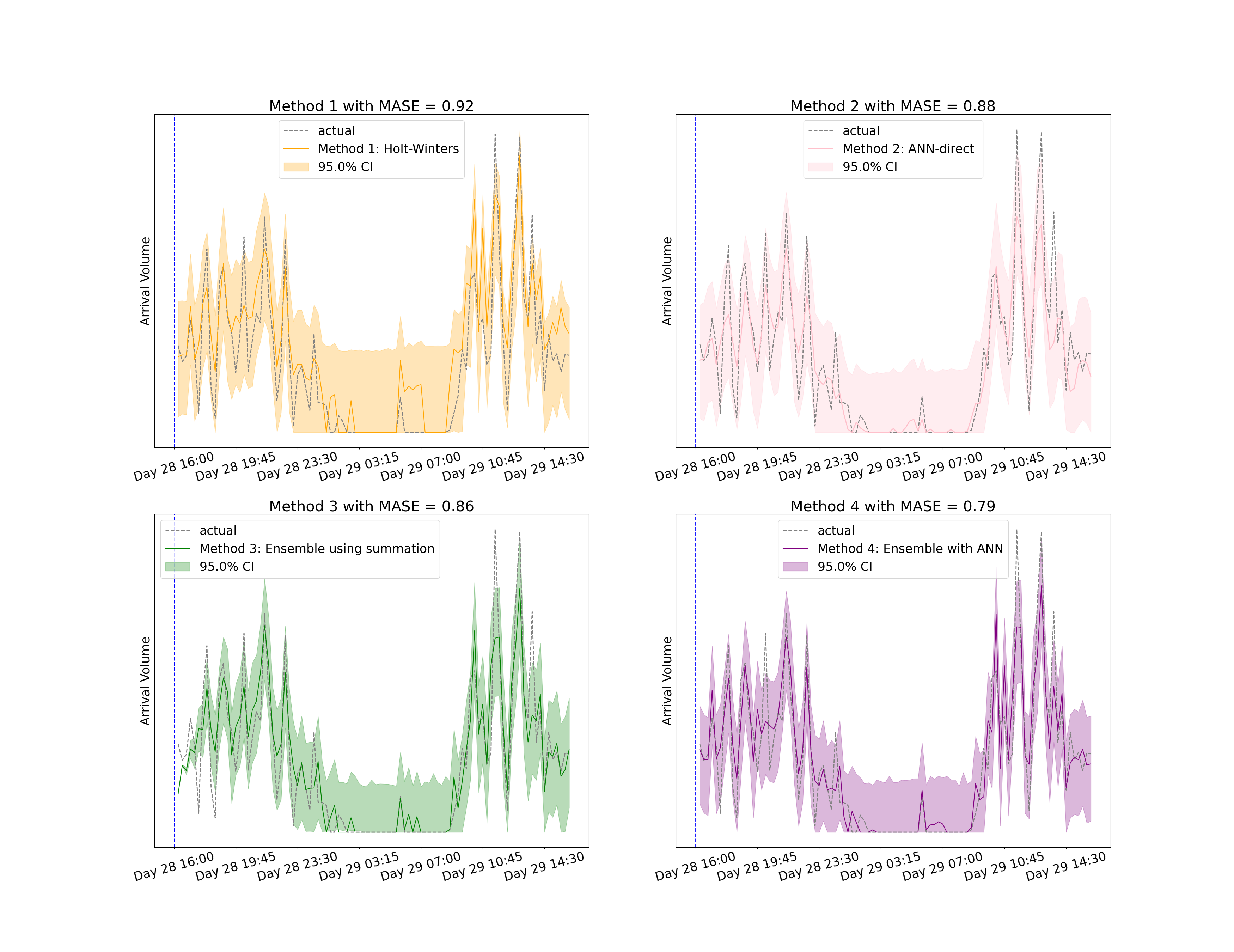}}

	\caption[Example of Forecasting Results]{Forecast results of all methods with the observation time at Day 28 16:00; Each of these four plots shares the same scale, with the X-axis representing the arrival time and the Y-axis representing the arrival volume.}
	\label{fig:fig9}
\end{figure}

In order to evaluate the performance of various forecasting methods, a time-series plot can be employed. Such plots provide a visual representation of the predicted outcomes, as well as any associated uncertainties. An illustrative example of this approach is presented in Figure \ref{fig:fig9}, which depicts the forecast results obtained at a specific observation time (Day 28, 16:00). It is worth noting that the shaded regions in Figure \ref{fig:fig9} correspond to the $95\%$ confidence intervals associated with each method's forecast. In time series forecasting, a confidence interval is a range of values around a forecasted point that is likely to contain the true value of the underlying time series variable with a certain level of confidence.

Of particular interest in this analysis is the width of these intervals, which serves as an indicator of the methods' respective degrees of uncertainty. Specifically, it can be observed that Method 1 exhibits the widest confidence interval, followed by Method 2. Additionally, our proposed method, Method 4, shows a confidence interval width that is similar to that of Method 3. A narrower confidence interval indicates greater confidence in the forecasted value, while a wider interval implies more uncertainty. Therefore, the smaller forecasting uncertainties observed in our proposed method provide additional evidence of its robustness and reliability in capturing and predicting the dynamics of parcel demand within the logistics network.

\section{Conclusion}
In this study, we tackled the parcel arrival forecast problem through three distinct tasks: forecasting the arrival of unordered parcels, predicting the arrival of ordered parcels, and creating an ensemble model that combines both forecasts using another neural network model. For the unordered parcel arrival forecast, we utilized a neural network model trained on historical data to predict the parcel arrivals $\hat{u}_t, t = 0, ..., T$ at any hub every $I$ minutes. For the ordered parcel prediction, we employed a random forest model trained on past data and updated parcel status to forecast the expected volume $\hat{o}_t, t = 0, ..., T$ of ordered parcels. By integrating these approaches with the ensemble model, we provided a more comprehensive and accurate parcel arrival forecast for the logistics network. Our ensemble method, which merges historical arrival patterns with live parcel status updates, enabled us to predict future workload more accurately, reducing uncertainty and facilitating more effective planning and resource allocation. Despite the limitations posed by a small dataset, our method demonstrated promising results, highlighting its potential for practical applications with larger datasets.

Overall, our research makes a significant contribution to logistics forecasting by introducing a dynamic, data-driven approach that can handle short-term, multi-step forecasts for parcel arrival volumes at logistics hubs. We demonstrated how integrating deep learning techniques with real-time data enhances prediction accuracy, showcasing strong potential for practical application even with limited data.

To further enhance this research, exploring additional data sources such as traffic conditions, weather forecasts, and social media trends could provide deeper insights and improve forecast accuracy by capturing external factors affecting parcel deliveries. Additionally, expanding the real-time data utilization capabilities of our models will allow for more dynamic adjustments to forecasts as new data becomes available, improving responsiveness and accuracy.

Looking toward scalability and deployment, future studies should investigate how our approach can be scaled and integrated into larger, more complex logistics networks. This includes assessing its effectiveness across different logistics hubs and potentially in other industries facing similar forecasting challenges, such as manufacturing or retail inventory management. Such studies will evaluate the versatility and adaptability of our forecasting techniques. By addressing these areas, future research can validate the effectiveness and reliability of our proposed model and enhance its practical applicability, ultimately aiding logistics companies in improving operational efficiency and making informed decisions for resource allocation and short-term planning.











\printcredits

\bibliographystyle{cas-model2-names}

\bibliography{cas-refs}

\end{document}